\documentclass[conference]{IEEEtran}
\IEEEoverridecommandlockouts
\usepackage{cite}
\usepackage{amsmath,amssymb,amsfonts}
\usepackage{graphicx}
\usepackage{textcomp}
\usepackage{xcolor}
\usepackage{caption}
\usepackage{subcaption}
\usepackage{float}
\usepackage{textcomp}
\usepackage{makeidx}
\usepackage{booktabs}
\usepackage{multicol}
\usepackage{algorithm}
\usepackage{fixltx2e}
\usepackage{stackengine}
\usepackage[noend]{algpseudocode}
\usepackage{url}
\usepackage{tablefootnote}

\def\BibTeX{{\rm B\kern-.05em{\sc i\kern-.025em b}\kern-.08em
    T\kern-.1667em\lower.7ex\hbox{E}\kern-.125emX}}
\begin{document}

\title{Multi Modal Semantic Segmentation using Synthetic Data \\
{\footnotesize }
\thanks{}
}
\author{\IEEEauthorblockN{
Kartik Srivastava\textsuperscript{*1},
Akash Kumar Singh\textsuperscript{*2},
Guruprasad M. Hegde\textsuperscript{3},
}

\IEEEauthorblockA{
    kartiksrivastava144@gmail.com,
    ksakash@iitk.ac.in,
    GuruprasadMahabaleshwar.Hegde@in.bosch.com
}
}

% \author{\IEEEauthorblockN{Kartik Srivastava\textsuperscript{*}}
% \IEEEauthorblockA{\textit{Department of Computer Science \& Information Systems} \\
% \textit{Birla Institute of Science and Technology, Pilani}\\
% Hyderabad, India \\
% kartiksrivastava144@gmail.com}

% \and
% \IEEEauthorblockN{Akash Kumar Singh\textsuperscript{*}}
% \IEEEauthorblockA{\textit{Department of Computer Science} \\
% \textit{Indian Institute of Technology, Kanpur}\\
% Kanpur, India \\
% ksakash@iitk.ac.in}

% \and
% \IEEEauthorblockN{Guruprasad M. Hegde}
% \IEEEauthorblockA{\textit{Robert Bosch Research and Technology Center} \\
% Bangalore, India\\
% GuruprasadMahabaleshwar.Hegde@in.bosch.com}
% }

\maketitle

\begin{abstract}
Semantic understanding of scenes in three-dimensional space (3D) is a quintessential part of robotics oriented applications such as autonomous driving as it provides geometric cues such as size, orientation and true distance of separation to objects which are crucial for taking mission critical decisions. As a first step, in this work we investigate the possibility of semantically classifying different parts of a given scene in 3D by learning the underlying geometric context in addition to the texture cues BUT in the absence of labelled real-world datasets. To this end we generate a large number of synthetic scenes, their pixel-wise labels and  corresponding 3D representations using CARLA software framework. We then build a deep neural network that learns underlying category specific 3D representation  and texture cues from color information of the rendered synthetic scenes. Further on we apply the learned model on different real world datasets to evaluate its performance. Our preliminary investigation of results show that  the neural network is able to learn the geometric context from synthetic scenes and effectively apply this knowledge to classify each point of a 3D representation of a scene in real-world.  

\newcommand\blfootnote[1]{%
  \begingroup
  \renewcommand\thefootnote{}\footnote{#1}%
  \addtocounter{footnote}{-1}%
  \endgroup
}
\blfootnote{\textsuperscript{1} \textit{Department of Computer Science \& Information Systems, \\Birla Institute of Technology and Science, Pilani}. Hyderabad, India}
\blfootnote{\textsuperscript{2} \textit{Department of Electrical Engineering, \\Indian Institute of Technology, Kanpur}. Kanpur, India.}
\blfootnote{\textsuperscript{3} \textit{Robert Bosch Research and Technology Center}. Bangalore, India.}
\blfootnote{\textsuperscript{*} Work done while employed as an intern at Robert Bosch Research and Technology Center, Bangalore.}
% Autonomous driving is slowly becoming one of the most exciting fields in research, as it could be a game-changing technology for future transport systems. With many industrial leaders such as Tesla, Uber and Google (Waymo) focusing their research into this field, it has become one of the most engaging problems of artificial intelligence.
% Recent methods for designing autonomous vehicles have focused on using machine learning techniques for both - controlling the vehicle as well as understanding the environment using various sensors.

% But a major problem with such methods is that they are data intensive. The state of the art deep learning based models requires thousands of instances of data to work well. The problem arises in case of non-ideal scenarios. Scenarios such as jaywalking, not adhering to traffic lights, obstacle avoidance and accidents are rarely available on real world datasets. And even if such scenarios are available, they are only isolated incidents recorded and are nowhere representative of the actual class scenarios.

% To tackle these shortcomings of publicly available real world datasets, we aim to create such scenarios using CARLA, which is an open-source simulator that has been developed to support development, training, and validation of autonomous driving systems.
\end{abstract}


\section{Introduction}\label{sec:intro}
% This section contains the intro to the problem and motivation for this work
 Semantic segmentation and understanding of a scene is one of the fundamental requirements for any robotics applications. The task of semantic segmentation involves classifying each element of a scene to a set of predefined categories/classes. For instance, in a point wise representation of a 3D world, the task involves classifying each point to a specific category. The role of semantic segmentation and challenges thereby is further emphasized in mission critical tasks such as self-driving technologies. For e.g. during path planning stage it is important to understand pose, size and distance to an object and make a decision whether to apply brakes or negotiate with the obstacle.

\begin{figure}
\begin{multicols}{2}
    \includegraphics[width=0.2\textwidth]{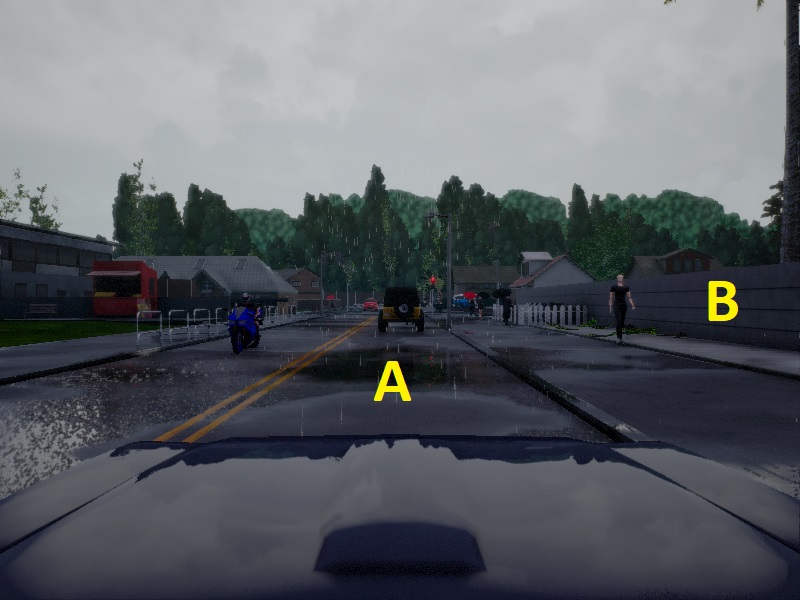}\par 
    \subcaption{}
    \includegraphics[width=\linewidth]{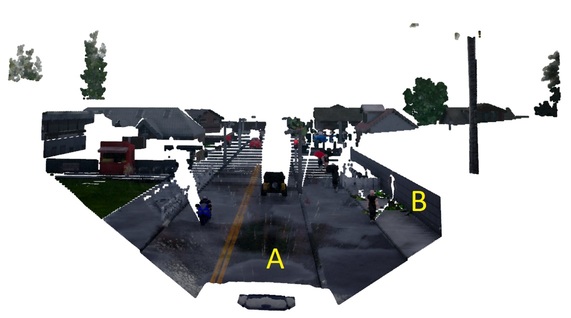}\par 
    \subcaption{}
    \end{multicols}
\begin{multicols}{2}
    \includegraphics[width=0.9\linewidth]{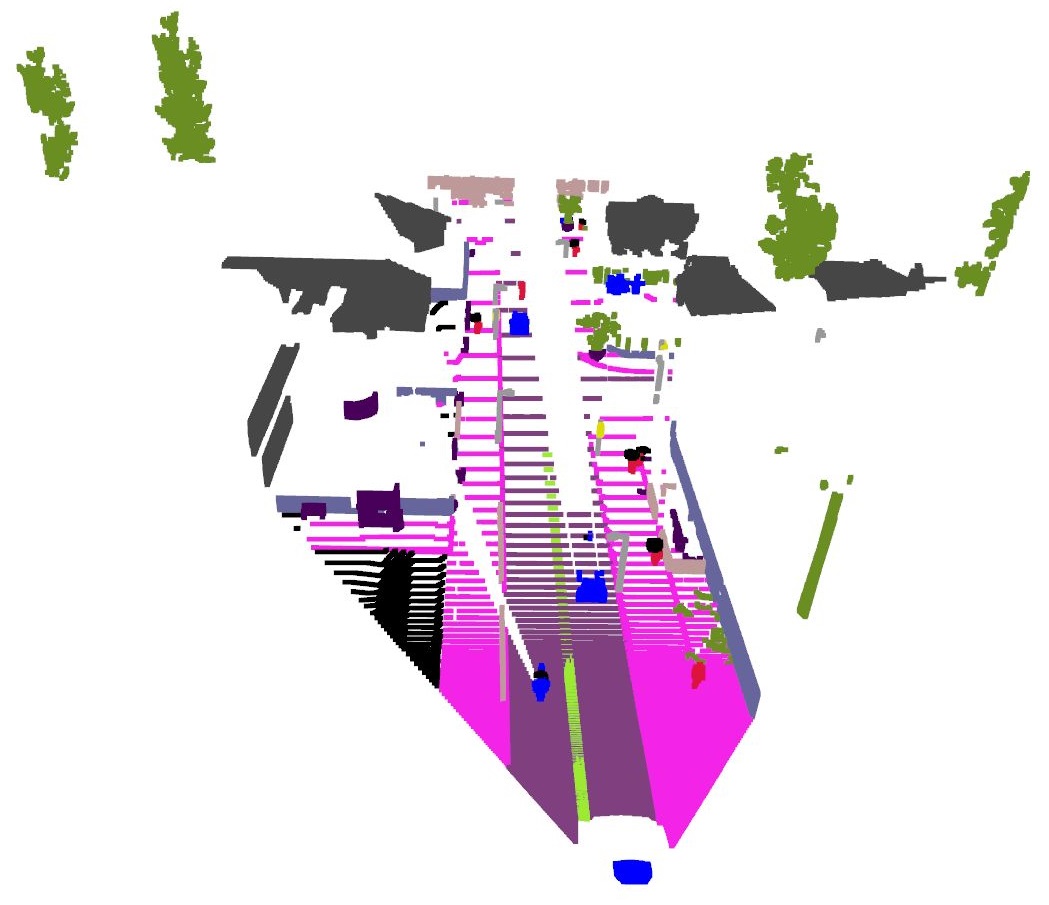}\par
    \subcaption{}
    \includegraphics[width=\linewidth]{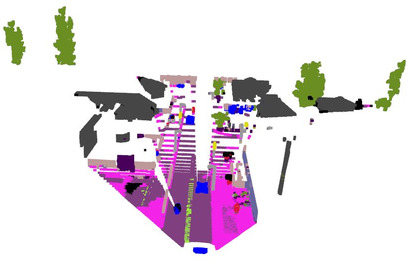}\par
    \subcaption{}
\end{multicols}
\caption{a)Rendered Synthetic image, b)Multi modal image consisting of 3D and color information, c)Ground Truth semantic labels, d) Our multi-modal semantic segmentation results }
\label{fig:carla_multi_view}
\end{figure}

 Research community has made tremendous progress in developing semantic segmentation methods in the two dimensional (2D) image space using Deep Neural Networks (DNN) \cite{semseg3}. However segmentation based on image information alone might not be sufficient for applications like self-driving since its difficult to obtain accurate 3D information such as pose, size, distance to an object to make critical decisions. In addition,  color information could be misleading as shown in Fig. \ref{fig:carla_multi_view}a which are synthetically generated overcast/rainy scene. We can observe both road (denoted by A) and wall (denoted by B) have similar color and might confuse the segmentation algorithm while assigning a class to a pixel.
 
 To address the above shortcomings of image based scene segmentation, self-driving technologies from companies like   Waymo\cite{waymo}, Lyft\cite{lyft} use multiple perception modules such as Light Detection And Ranging (LiDAR) devices in addition to cameras. LiDARs work based on active illumination and represent a scene in discrete set of points in 3D, also known as a Point Clouds (PC). For instance, in Fig. \ref{fig:carla_multi_view}b the LiDAR information is fused with color information which helps to distinguish horizontal road and vertical wall based on their distinct orientation in 3D, resulting in better segmenting as shown in Fig. \ref{fig:carla_multi_view}d. The corresponding Ground Truth (GT) labels are shown in Fig. \ref{fig:carla_multi_view}c where each class is represented by a distinct color (e.g. Road is represented by dark pink and Wall is represented by Gray). Based on the above factors it is important to develop and extend existing techniques to handle 3D representation of a scene such as PC inputs in addition to images. Currently there is far less work on how to adapt 2D image based segmentation techniques to  3D point clouds. One of the major bottlenecks being non-availability of labelled real world PC datasets.

% Large scale publicly available point cloud datasets  like Waymo\cite{waymo}, Nuscenes\cite{nuscenes} provide coarse labels in the form of bounding boxes and not sufficient in case of semantic segmentation which require point-wise labels. One of the reason for this is the labelling complexity and thereby huge cost to manually annotate each individual point in the 3D scan. A representative 360 degree view of a single LiDAR scan and corresponding Ground Truth (GT) labels are shown in Fig.\ref{fig:raw_3d} where each color represents a class - e.g. blue represents cars, pink represents side-walk etc. In the absence of color information it is unnatural and difficult for a human to establish correspondence between points and object class they belong to, resulting in a complex and arduous labelling process. \cite{sem-kitti}.  

% Another problem with real world datasets is the scarcity of various rare scenarios. For instance its difficult to capture scenarios of  collision between vehicles, collision due to jay-walkers, negotiating with cars on wrong way, pre-emptive breaking to avoid vehicles running over the stop signal etc. It is extremely important to capture such non-ideal scenarios experienced in the real world to train and test performance of semantic segmentation, which determines the practicality of developed techniques. 

Considering the above mentioned factors, in this work we investigate the usability of computer generated labelled synthetic datasets for semantic segmentation of PC. Along this direction our contributions include:
\begin{itemize}
    \item  Developing a run-time and memory efficient implementation of state of art PC segmentation algorithm --- Pointnet++\cite{pointnet++} and is described in section \ref{subsec:Runtime and memomry optimizations} 
    \item Validating the effectiveness of resulting segmentation algorithm to learn geometric cues from synthetic datasets and transferring the learnt knowledge to real world datasets as presented in section \ref{sec:exp_and_results}
\end{itemize}

% two fold, firstly, we modify an existing state of art PC segmentation algorithm to make it run-time and memory efficient and Second, we validate the effectiveness of resulting segmentation algorithm to learn geometric cues from synthetic datasets and transferring the learnt knowledge to real world datasets. 

In this paper we first briefly review some of the publicly available labelled 3D point cloud datasets and methods in 3D semantic segmentation in section \ref{sec:rel_work}, followed by problem description and proposed solution and contributions in section \ref{sec:prob_statement}. We deliberate on experiments and results in section \ref{sec:exp_and_results} and finally conclude in section \ref{sec:conclusion_and_future_work}. 

%  in section \ref{sec:carla} we introduce CARLA simulator which we have used to generate synthetic data. Section \ref{sec:prob_statement} briefly describes our problem statement followed by our proposed solution in section \ref{sec:proposed_soln}. We deliberate on experiments and results in section \ref{sec:exp_and_results} and finally conclude and discuss future directions in section \ref{sec:conclusion_and_future_work}

\section{Related Work}\label{sec:rel_work}
% This section outlines some related work in this direction.
\subsection{Datasets for 3D semantic segmentation}
% Many real world datasets and few synthetic datasets have been released for semantic segmentation of 3D point clouds, such as the Oakland data set \cite{oakland3d}, NYU Depth Dataset V2 \cite{nyu3d}, Sydney Urban Objects data set \cite{sydney3d}, IQmulus \& TerraMobilita Contest \cite{terra3d},  Vaihingen3D airborne benchmark \cite{Vaihingen3D}, Paris-lille 3D dataset \cite{},  Semantic-Kitti \cite{sem-kitti} and Apolloscapes \cite{apolloscapes},virtual kitti \cite{virtual-kitti} ,Synth-city \cite{synth-city}.

There are a few available labelled datasets \cite{oakland3d}\cite{sem3d}\cite{apolloscapes}, but they suffer from lack of sufficient representation of relevant classes for self driving applications \cite{sem3d}\cite{apolloscapes} or consist of few labelled points\cite{oakland3d}. The most recent Sem-kitti \cite{sem-kitti} and Synth-city \cite{synth-city} datasets are promising but suffer from unavailability of labelled data for rare events (e.g. collision between vehicles) which are crucial to evaluate the practicality of developed techniques in the context of self-driving. 
%   Few outdoor real world \cite{oakland3d}\cite{sem3d}\cite{sem-kitti}\cite{apolloscapes} and synthetic labelled datasets \cite{virtual-kitti}\cite{synth-city} are available for semantic segmentation of 3D point clouds. Most of these datasets either have few labelled points\cite{oakland3d}, few labelled classes that are relevant in the context of self-driving\cite{apolloscapes}, or  might not truly represent the sensor characteristics of a LiDAR intended for self-driving applications\cite{sem3d}. The more recently released  semantic-kitti dataset \cite{sem-kitti} provides labels for around 23401 LiDAR scans which is relevant and we will be investigating in our future work. 

%  In the realm of synthetic datasets, virtual kitti\cite{virtual-kitti} provides labels for 17000 scenes but does not include LiDAR scans. Recently released Synth-city\cite{synth-city} dataset is based on realistic rendering of an automotive grade LiDAR and needs to be investigated further. Although there is potential in using many of the above datasets they still fall short of providing labelled data on rear events such as collision between vehicles , Jaywalkers running over a stop signal etc. which is essential to understand performance of developed semantic segmentation techniques on unseen configurations. Hence it is important to investigate a simulation platform that renders a realistic operating environment of a self-driving car and also allows the freedom to create scenarios according to user\textquotesingle s choice. 
 
 To study the usability of synthetic data for semantic segmentation of real world point clouds, in this work we test our approaches predominantly on KITTI dataset \cite{kitti_dataset}  which is closer to our use case and will be described in section \ref{sec:exp_and_results}. 

\subsection{Segmentation of Point clouds}\label{subsec:relwork_segpcd}
Semantic segmentation of 3D point clouds has been one of the most well researched problem. Related techniques involve non-parametric approaches based on region growing \cite{3dseg1}, segmenting based on graph partitioning \cite{3dseg2}\cite{3dseg5}, robust statistics to fit a pre-defined parametric model \cite{3dseg3}\cite{3dseg4}. Most of these methods perform well on constrained data sets with very limited variation in operating conditions. With the advent of large scale image data sets with high variations and the potential shown by DNN in semantically segmenting them \cite{semseg3}, it is natural that most of the recent point cloud segmentation methods \cite{voxel}\cite{pointnet++}\cite{squeezeseg} are exploring similar ideas. 

Most of these techniques initially transform a 3D PC to a suitable representation like Voxel grid  \cite{voxel} or to images using projection techniques\cite{squeezeseg} and then apply well studied methods based on DNN for images. Since the processing is performed on a derived representation of raw point clouds which incurs loss of information, it might not be the most effective way to understand the benefits of synthetic data. For this reason we investigate  Pointnet++ \cite{pointnet++} which  directly process raw 3D point clouds and will be discussed in \ref{sec:prob_statement}.

\section{Problem statement and Solution}\label{sec:prob_statement}
% This section contains the formal problem statement
 As discussed in section \ref{subsec:relwork_segpcd} it is well proven that fully supervised techniques based on DNN are ideal for semantic segmentation of images and some of the early work show potential by extending these techniques to segment 3D point clouds. However there are three major bottlenecks which hinders the progress of research in the field of 3D PC segmentation. 1) Sensor setup to acquire 3D PC that are suitable for self-driving applications is extremely expensive. 2) Complexity and thereby huge costs involved in manual labelling of PC and 3) which partly is a consequence of 1) and 2) is the scarcity of real-world labelled PC data in both normal and rare events. In the following section we propose our solution to address the listed problems.  
 
% \section{Proposed solution}\label{sec:proposed_soln}
To overcome the problem of scarcity of labelled data for normal and rare events we propose to leverage  synthetic data. The synthetic data generator should represent a realistic operating environment of self-driving vehicle and should be able to provide labels for the underlying classes. It should also provide a programming interface to  enable generation of rare events such as vehicle collisions. To fulfill the above conditions we leverage the CARLA simulator\cite{carla}. The detailed description of CARLA is provided in section \ref{subsec:carla}.

To reduce efforts in labelling of real-world 3D PC we propose a coarse to fine approach. The coarse labelling is provided by initially learning 3D representations from synthetic data using DNN that directly works on 3D points. The learnt model is then applied on real-world 3D PC to obtain coarse labels. The coarse labels could then be further refined or corrected manually. This could potentially save huge amounts of manual labor and consequently minimize cost of labelling. The details of DNN employed in this work is provided in section \ref{subsec:propsoln_pointnet}

% \subsection{Generating labelled point clouds using CARLA simulator}
% As mentioned in section \ref{subsec:carla_sensors} CARLA renders a velodyne 64 beam laser scanner along with semantic labels for images generated from color camera  that is mounted on a ego-vehicle. We use this setting to create a dataset of 4000 scenes (having over 1.6 billion points) with 16 different weather conditions spread across two different towns. All the scenes has varying traffic and pedestrian density, and the data was collected in urban and sub-urban settings. The dataset also consists of 3D PC related to generated scenarios as shown in section \ref{subsec:carla_scene_gen}.
\subsection{Synthetic data generation}\label{subsec:carla}
% \section{CARLA Simulator }\label{sec:carla}
\subsubsection{Sensor suite }\label{subsubsec:carla_sensors}
\begin{figure}[h!]
    \centering
    \includegraphics[width=\linewidth]{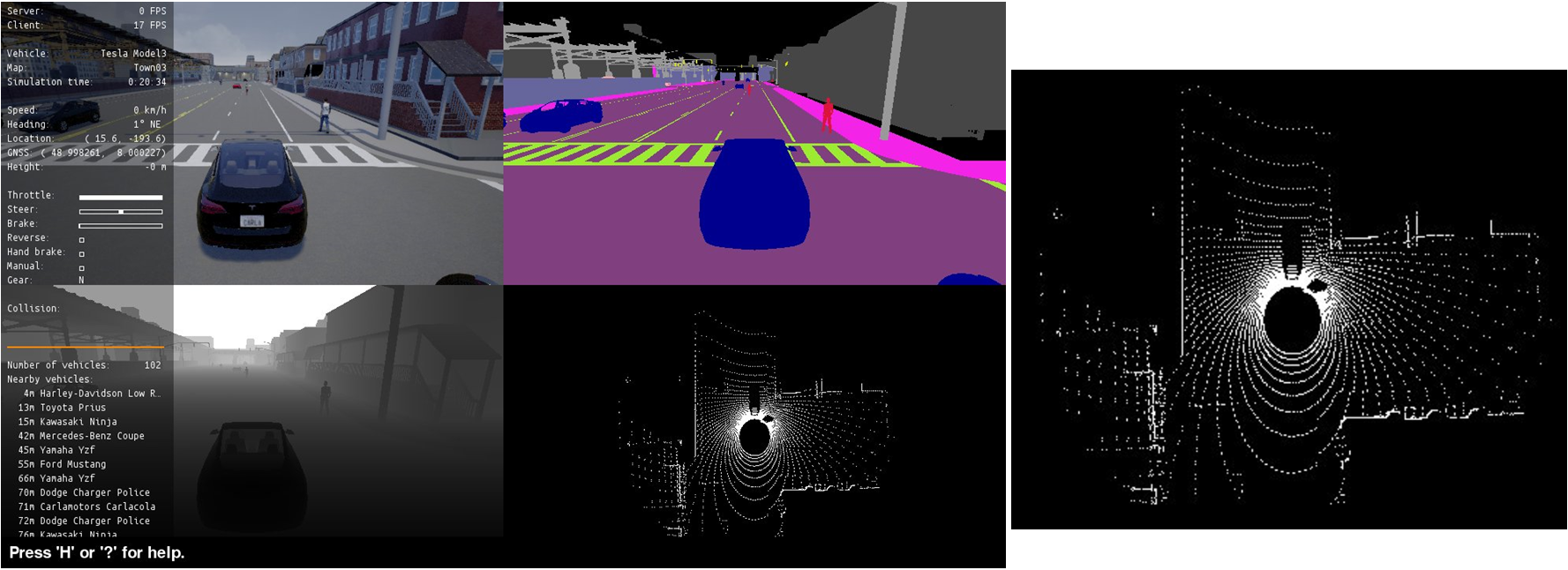}
    \caption{A snapshot of sensors rendered in CARLA. The available sensor information are Color image, Semantic labels of 2D image, LiDAR and depth camera (from top-left in clockwise manner). Zoomed view of LiDAR is shown in the right image}
    \label{fig:multi_view}
\end{figure}

In this work we use CARLA\cite{carla}, an open source driving simulator for self-driving research to generate synthetic data. CARLA simulates  commonly used sensors in autonomous driving such as cameras and LiDARs as shown in Fig.\ref{fig:multi_view}. 

We fuse the point clouds with color information and semantic labels to obtain labelled 3D PC. The details of generated dataset will be provided in section \ref{subsec:Datasets}.

% \begin{figure*}[!ht]
%      \begin{minipage}[l]{1.0\columnwidth}
%          \centering
%          \includegraphics[width=\linewidth]{images/carla_training_data/Picture_1.png}
%     \label{fig:train_data_!}
%      \end{minipage}
%      \hfill{}
%      \begin{minipage}[r]{1.0\columnwidth}
%          \centering
%          \includegraphics[width=\linewidth]{images/carla_training_data/Picture_2.png}
%          \label{fig:train_data_2}
%      \end{minipage}
%      \caption{Samples from the training data generated by CARLA simulator}
% \label{fig:sim_2d}
% \end{figure*}

% Some of the sample images generated using CARLA is shown in Fig.\ref{fig:sim_2d}. Corresponding 3D PC with labels were used for training the semantic segmentation model.

\subsection{Scenario Generation using CARLA simulator} \label{subsec:carla_scene_gen}
\begin{figure}[h!]
    \centering
    \begin{subfigure}[h!]{\linewidth}
        \centering
        \includegraphics[width=0.45\textwidth]{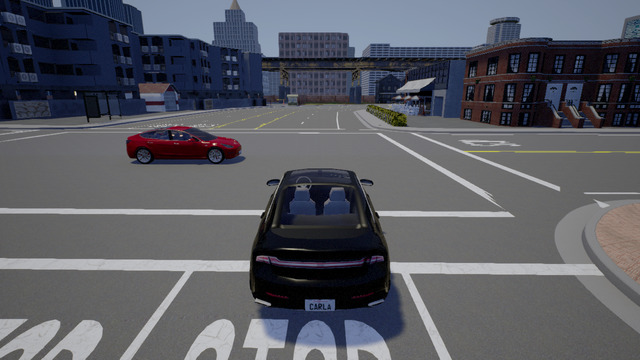}
        \includegraphics[width=0.45\textwidth]{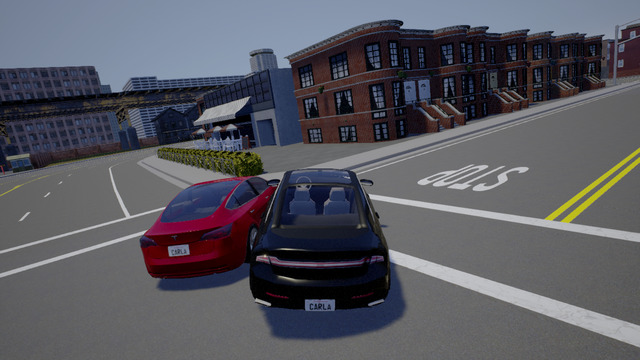}
        % \caption{}
        \label{fig:scenario_crash}
    \end{subfigure}
    
    % \begin{subfigure}[h!]{\linewidth}
    %     \centering
    %     \includegraphics[width=0.45\textwidth]{images/scenarios/Jaywalker/00058565.jpg}
    %     \includegraphics[width=0.45\textwidth]{images/scenarios/Jaywalker/00058569.jpg}
    %     \caption{}
    %     \label{fig:scenario_jaywalkwer}
    % \end{subfigure}
    % \caption{a)Generated scenario of car crashing into another, b)Jaywalker crossing the road when the signal is turned Green}
    % \label{fig:scenario_generation}
     \caption{Generated scenario of two cars crashing into each other}
    \label{fig:scenario_generation}
\end{figure}

Rare scenarios such as accidents and near misses do not appear in datasets in abundance in real world. To address this gap we generate scenarios in CARLA using the provided application programming interface and include them as part of our training and validation set. An example scenario of two cars crashing each other is shown in Fig. \ref{fig:scenario_generation}

The simulator offers flexibility to configure different weather conditions, actor (e.g. pedestrians, vehicles) density and town layouts. We use these configuration parameters to create a diverse set of synthetic dataset. More specifically we performed the following procedure:
\begin{itemize}
    \item Define the actors in the scenario, along with their desired behaviour (run in a straight line or autopilot motion) and location of spawning in the CARLA world.
    \item Define the ego-vehicle along with it\textquotesingle s initial spawning point and sensors for collecting data.
    \item Define the behaviour of the actors and when to trigger them with respect to the location of the ego-vehicle.
    \item Create the CARLA server and launch the actors and ego-vehicles in the simulation.
    \item Maneuver the ego-vehicle along the desired path and capture the scenario of how the actors respond to the ego-vehicle by dumping the data from the sensors.
\end{itemize}

% \begin{figure}[h!]
%     \centering
%     \begin{subfigure}[h!]{\linewidth}
%         \centering
%         \includegraphics[width=0.9\textwidth]{images/kitti_2d.jpg}
%         \caption{}
%         \label{fig:kitti_2d}
%     \end{subfigure}
    
%     \begin{subfigure}[h!]{\linewidth}
%         \centering
%         \includegraphics[width=0.9\textwidth]{images/kitti_3d.jpg}
%         \caption{}
%         \label{fig:kitti_3d}
%     \end{subfigure}
    
%     \begin{subfigure}[h!]{\linewidth}
%         \centering
%         \includegraphics[width=0.9\textwidth]{images/kitti_3d_semseg.jpg}
%         \caption{}
%         \label{fig:kitti_3d_semseg}
%     \end{subfigure}
%     \caption{a)Sample 2D image from KITTI dataset, b)Colored point cloud of a scene from KITTI dataset obtained by registration of RGB image with point cloud, c)Labelled point cloud of a scene from KITTI dataset obtained by registration of semantic segmentation image with point cloud}
% \label{fig:kitti_example}
% \end{figure}
\subsection{Deep Neural Network for point cloud segmentation }\label{subsec:propsoln_pointnet}
In this work we base our investigation on Open3D implementation of  PointNet++\cite{o3d_pointnet} which is one of the popular DNN architecture that takes point clouds as input and predicts class labels for each individual point.
% The main building blocks are:

% % \begin{figure}[h!]
% %     \centering
% %     \includegraphics[width=\linewidth]{images/pointnet_model.png}
% %     \caption{The PointNet model architecture}
% %     \label{fig:pointnet}
% % \end{figure}

% \begin{itemize}
%     \item \textbf{Local Feature extractor:} In this stage Pointnets\cite{pointnet} is applied to a local neighborhood of each point that forms the feature vector. 
%     \item \textbf{Global Feature extractor:} The output from local feature extractor is combined across multiples scales that enables capturing complex global context such as, surface geometry, spatial relationships between different structure in the scene etc. 
%     \item \textbf{Classifier:} This block consists of a series of fully connected layers that take the combined local and global feature as input and provides a score for each class as output.  
% \end{itemize}

 A typical LiDAR point cloud consists of a few million 3D points. Considering the fact that these data points needs to be represented as data type double, the required memory footprint to load the whole data-set at once, for the purpose of training a DNN becomes impractical and sluggish assuming a standard desktop RAM of 8GB. Even in case of batch-wise loading from a secondary memory like hard drive the process of reading every batch sequentially while training is too time consuming. To address this gap in default implementation, we developed an efficient way to generate and load batches of data as described below.
% brief description of pointnets
% \subsection{implementation details of multi-processing}
% why this is needed and what benefits it brings in to the stock implementation
% \subsection{implementation details of model initialisation }

%---------------- added ------------------%
\subsection{Runtime and memomry optimizations}\label{subsec:Runtime and memomry optimizations}
% \begin{itemize}
%     \item \textbf{Efficient Batch Creation and Loading} : 
    
    % \hspace{4mm} We address this problem will be to not load the whole dataset at once but to read the data file directly from the secondary memory one by one which doesn\textquotesingle t possess any threat on the RAM size limit. But the main problem with this approach is the time that goes into reading a file from secondary memory. So it adversely affect the training time for a model. Also in this approach we are not making use of the multiple cores provided by the CPU, which can help in improving run time of batch creation and loading.

    % \hspace{4mm} In our approach we modified the open3d implementation of Pointnet++ \cite{o3d_pointnet} to handle large point clouds and for better runtime performance. The main idea is to store very limited data on RAM which are accessed from a parallel process at the time of training for generating and feeding continuous batches of data. The pseudo code of our method is shown in Algorithm 1 below.
    \hspace{4mm} The main idea is to store very limited data on RAM which are accessed from a parallel process at the time of training for generating and feeding continuous batches of data. The pseudo code of our method is shown in Algorithm 1 below.
    % Firstly we store very limited data on RAM such as  filename, no. of points and location of data points. Second, we  parallelize the batch generation and loading process that harness multiple CPU cores.

    % \hspace{4mm} To reduce frequent use of RAM we created a special class to handle information such as dataset location, no. of files in each point cloud and methods to load labels, create batch, provide different option for the data, etc.\par

    %  Second we create a parallel process beside the main training process in the script. This new process is dedicated only for creating batches and loading it into RAM.
    
    \begin{algorithm}
    \caption{Training Data Stacking Algorithm}\label{basic}
    \begin{algorithmic}[1]  
        \Statex \textbf{Result:} \textit{Queue} : queue to store batches for training
    	\State \textbf{Initialize:}
    	\State \textit{Queue} : queue to store batches for training
    	\State \textit{QueueLimit} : limit on the size of the queue
    	\State \textit{Buffer} : temp buffer before filling the queue
    	\State \textit{BufferLimit} : limit on the size of buffer
        \State \textit{Dataset} : class designed to manage the dataset
    	\State \textbf{FillQueue} : 
    	\While{\textit{True}}
    		\If {\textbf{len}(\textit{Buffer}) \textless \hspace{2pt} \textit{BufferLimit}}  
    			\State \textit{Buffer}\textbf{.} \textit{Append}(\textbf{GetBatch})
    		\EndIf
    		\For{\textit{P} \textbf{in} \textit{Buffer}}
                \If {\textbf{len}(\textit{Queue}) $>=$ \textit{QueueLimit}}
                    \State \textbf{return}    
                \EndIf
            \State \textit{Queue}\textbf{.} \textit{Put}(\textit{P})
            \State \textit{Buffer}\textbf{.} \textit{Remove}(\textit{P})
            \EndFor
    		\State \textit{time.sleep(0.01)}
    	\EndWhile
        \State \textbf{GetBatch: }
        \State \textbf{return} \textit{Dataset}\textbf{.} \textit{SampleBatch}
    \end{algorithmic}
    \label{train-stack}
    \end{algorithm}
% \end{itemize}
    \hspace{4mm} The \textbf{FillQueue} method runs in parallel with the training script to continuously feed generated batches of data to the \textit{Queue} which resides on RAM. We first initialize an array to contain batches of training data. This acts as a buffer for \textit{Queue}. Next, the data is appended to \textit{Buffer}. Subsequently the data in the \textit{Buffer} is pushed to the \textit{Queue} and corresponding elements are cleared from the \textit{Buffer}. BufferLimit and queue limit are two parameters that limit the size of data stored on RAM so that memory foot print can be controlled. \textbf{GetBatch} is the method which generates batches of data from secondary memory using information on RAM such as filename, no. of points and location of data points which belongs to the \textit{Dataset} class.
    
    By limiting the amount of data residing on RAM and on-demand creation of batches of data in a parallel manner makes the training process  run-time efficient.

%------------DEFINING SEMANTIC LABEL COLORS-----------%
%----------------------------------------------------%
\definecolor{Building}{RGB}{70,70, 70}
\definecolor{Road}{RGB}{128,64,128}
\definecolor{Side-walk}{RGB}{244,35,232}
\definecolor{Vegetation}{RGB}{107,142,35}
\definecolor{Car}{RGB}{0,0,255}
\definecolor{Fence}{RGB}{190,153,153}

\definecolor{Cityscapes_Road}{RGB}{128, 64, 128}
\definecolor{Cityscapes_Sidewalk}{RGB}{244, 35, 232}
\definecolor{Cityscapes_Building}{RGB}{70, 70, 70}
\definecolor{Cityscapes_Wall}{RGB}{102, 102, 156}
\definecolor{Cityscapes_Fence}{RGB}{190, 153, 153}
\definecolor{Cityscapes_Pole}{RGB}{153, 153, 153}
\definecolor{Cityscapes_Traffic_Light}{RGB}{250, 170, 30}
\definecolor{Cityscapes_Traffic_Sign}{RGB}{220, 220,  0}
\definecolor{Cityscapes_Vegetation}{RGB}{107, 142, 35}
\definecolor{Cityscapes_Terrain}{RGB}{152, 251, 152}
\definecolor{Cityscapes_Sky}{RGB}{70, 130, 180}
\definecolor{Cityscapes_Person}{RGB}{220, 20, 60}
\definecolor{Cityscapes_Rider}{RGB}{255, 0,  0}
\definecolor{Cityscapes_Car}{RGB}{0,  0, 142}
\definecolor{Cityscapes_Truck}{RGB}{0,  0,  70}
\definecolor{Cityscapes_Bus}{RGB}{0, 60, 100}
\definecolor{Cityscapes_Train}{RGB}{0, 80,100}
\definecolor{Cityscapes_Motorcycle}{RGB}{0,  0, 230}
\definecolor{Cityscapes_Bicycle}{RGB}{119, 11, 32}

\definecolor{CARLA_Unlabeled}{RGB}{0, 0, 0}
\definecolor{CARLA_Building}{RGB}{70, 70, 70}
\definecolor{CARLA_Fence}{RGB}{190, 153, 153}
\definecolor{CARLA_Other}{RGB}{72, 0, 90}
\definecolor{CARLA_Pedestrian}{RGB}{220, 20, 60}
\definecolor{CARLA_Pole}{RGB}{153, 153, 153}
\definecolor{CARLA_Road_line}{RGB}{157, 234, 50}
\definecolor{CARLA_Road}{RGB}{128, 64, 128}
\definecolor{CARLA_Sidewalk}{RGB}{244, 35, 232}
\definecolor{CARLA_Vegetation}{RGB}{107, 142, 35}
\definecolor{CARLA_Car}{RGB}{0, 0, 255}
\definecolor{CARLA_Wall}{RGB}{102, 102, 156}
\definecolor{CARLA_Traffic_sign}{RGB}{220, 220, 0}

\definecolor{sem3d_unlabelled}{RGB}{0, 0, 0}
\definecolor{sem3d_man_made_terrain}{RGB}{0, 0, 255}
\definecolor{sem3d_natural_terrain}{RGB}{128, 0, 0}
\definecolor{sem3d_high_vegetation}{RGB}{255, 0, 255}
\definecolor{sem3d_low_vegetation}{RGB}{0, 128, 0}
\definecolor{sem3d_buildings}{RGB}{255, 0, 0}
\definecolor{sem3d_hard_scape}{RGB}{128, 0, 128}
\definecolor{sem3d_scanning_artefacts}{RGB}{0, 0, 128}
\definecolor{sem3d_cars}{RGB}{128, 128, 0}
%----------------------------------------------------%
%----------------------------------------------------%

\section{Experiments and Results}\label{sec:exp_and_results}
\subsection{Datasets}\label{subsec:Datasets}
We have used three datasets for our experiments - Semantic 3D\cite{sem3d}, KITTI\cite{kitti_dataset} and in-house generated dataset using CARLA simulator. Each of these datasets are unique with respect to number of scenes captured, total number of 3D points, number of classes and type of sensor used while data acquisition. The above aspects are summarized in Table \ref{tab_datasets}

\begin{table}[h!]
\caption{Datasets used for Experiments}
\begin{center}
\begin{tabular}{|c|c|c|c|c|}
\hline
\textbf{Dataset} & \textbf{\textit{Semantic}}& \textbf{\textit{Size}}& \textbf{\textit{Total Number}} &\textbf{\textit{Sensor}} \\
\textbf{} & \textbf{\textit{Classes}}& \textbf{\textit{}}& \textbf{\textit{of Points}} & \\
\hline
Semantic 3D & 8 & 30 scenes & 4 billion & TLS\\
KITTI & 19 & 93 scenes & 1.6 million & LiDAR\\
CARLA & 12 & 4400 scenes & 1.6 billion & Simulation\\
\hline
\multicolumn{5}{l}{TLS: Terrestrial Laser Scanner, LiDAR:  Light Detection And Ranging}
\end{tabular}
\label{tab_datasets}
\end{center}
\end{table}

\begin{figure}[h!]
    \centering
    
        \centering
        \includegraphics[width=\linewidth]{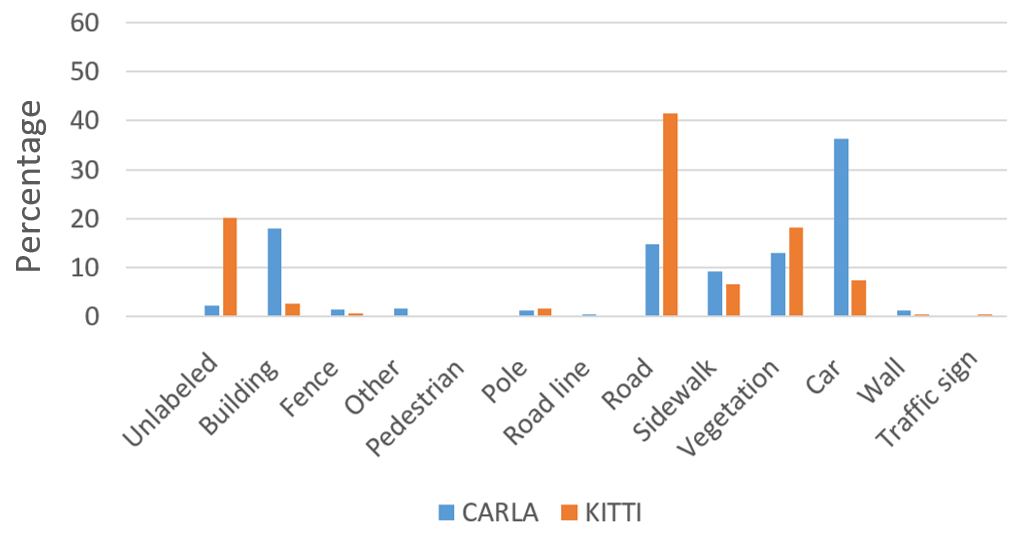}
        \caption{Classwise distribution of labelled points}
        \label{fig:class_dist}
    
\end{figure}
We used a standard split of 80\%-20\% for training and validating the Pointnet++ method on Semantic 3D and CARLA datasets. We used KITTI for testing the effectiveness of trained models using synthetic data generated by CARLA. We register the semantic GT with corresponding PC to form the labelled 3D representation prior to training and evaluation. An example of GT semantic labels are shown in  Fig. \ref{fig:expB}b and corresponding 3D labelled PC is shown in Fig. \ref{fig:expB}c.

 For diverse data acquisition in CARLA we spawned ego vehicle mounted with single camera and LiDAR from different locations and drove it on autopilot to cruise through the provided urban and semi-urban town environments. The traffic density was varied between 40 to 80 vehicles and number of pedestrians varied between 10 and 30. In addition, we cycled through 14 different weather conditions. In total we generated a set of 4000 scenes (having over 1.6 billion points).  The dataset also consists of 3D PC corresponding to generated rare scenarios as shown in section \ref{subsec:carla_scene_gen}. Considering each time an ego vehicle is spawned at a different location,  with varying vehicle and pedestrian density  along with change in weather conditions and town settings we ensure that the generated dataset provides a diverse training and validation set. 

Since each of the mentioned datasets have different number of semantic classes (Semantic 3D has 8 semantic classes, KITTI has 19 semantic classes and CARLA has 12 semantic classes) it is important to understand the classwise distribution to support quantitave evaluations. We show the classwise distribution of KITTI and CARLA in Fig. \ref{fig:class_dist} as both of them represent similar operating environment. There are five dominant classes in the two datasets namely Building, Road, Side-walk, Vegetation and Car. Since side-walk class is not present in Semantic-3D we restrict our evaluations to remaining 4 classes and mark the side-walk as unlabelled.  The re-mapping of Semantic-3D and KITTI classes to CARLA is shown in Table \ref{table:sem_class_map_all_to_common4}

\begin{table}[h!]
\caption{Mapping of semantic classes of different datasets to common 4 classes}
\begin{center}
\begin{tabular}{|c|c|c|c|}
\hline
\textbf{Common 4} & \textbf{Semantic-3D\textsuperscript{*}} & \textbf{CARLA\textsuperscript{*}} & \textbf{KITTI\textsuperscript{*}}\\
\hline
\textbf{Building} & Building & Building & Building\\
\hline
\textbf{Road} & Man-made Terrain & Road, Road-line & Road\\
\hline
\textbf{Car} & Car & Car & Car, Motorcycle,\\
& & & Bus, Bicycle\\
\hline
\textbf{Vegetation} & High-vegetation, & Vegetation & Vegetation\\
&  Low-vegetation & & \\
\hline
\multicolumn{4}{l}{$^{\mathrm{*}}$All remaining classes were mapped to unlabelled.}

\end{tabular}
\label{table:sem_class_map_all_to_common4}
\end{center}
\end{table}

\subsection{Evaluation metrics and criteria}
To assess labeling performance,  we used the standard Jaccard Index, commonly known as the PASCAL VOC intersection-over-union metric (IoU)\cite{iou}. To evaluate overall performance of our model, we used the standard mean intersection-over-union metric ($mIoU$), which was calculated by taking the average of IoU values of individual classes.

\begin{equation}\label{eqn:iou}
    IoU = \frac{TP}{TP+FP+FN}
\end{equation}

\begin{equation}\label{eqn:miou}
    mIoU = \frac{\sum_{i=1}^{n} IoU_{i}}{n}
\end{equation}

where where $TP$, $FP$, and $FN$ are  the  numbers  of  true  positive,  false  positive,  and  false negative pixels, respectively and $n$ is the number of semantic classes of the dataset.
\subsection{Training setup}
As mentioned in section \ref{sec:prob_statement} we use our modified version of Pointnet++ for training and testing the effectiveness of synthetic data. We train the model  on point clouds with and without color information denoted as RGB-D\textsubscript{\textit{i}-*} and D\textsubscript{\textit{i}-*} respectively. Here * represents the type of training dataset, \textit{i} represents the number of training classes (excluding unlabelled class), RGB stands for Red, Green, Blue channel of image and  D stands for Depth information. We trained all models until saturation, and used Adam optimizer with learning rate of $0.001$, momentum of $0.9$ and learning rate decay of $0.7$ . All models are trained on a sample size of $8192$ points per batch. 
\subsection{Results}
We trained and tested the Pointnet++ model on Semantic-3D dataset to establish a baseline on a real-world dataset. The results are captured in second row of Table \ref{tab2}. We attribute the high performance of the model mainly due to large population of labelled points for the 4 classes and the quality of static LiDAR scans which is less noisy. Further on we trained the models using the CARLA dataset and tested on Semantic-3D and KITTI datasets. Our aim is to understand how close we can match the baseline performance by training only on the synthetic dataset. The results are captured in rows three and four of Table \ref{tab2}. 

As expected the model performs well on Semantic-3D due to very high quality of TLS. Also the contribution from color information is minimal which is reflected in less difference in mIOU values of RGB-D and D-model. One of the possible reason cold be that majority of the classes have well defined geometry that is quite distinctive. On the KITTI dataset the performance of model trained on CARLA is lower and can be attributed to the relatively lower quality of LiDAR scan and resulting artifacts due to motion of the ego-vehicle. 
\begin{table}[h!]
\caption{Experimental comparison on 4 semantic classes}
\begin{center}
\begin{tabular}{|c|c|c|c|c|c|}
\hline
\textbf{Training} & \textbf{Validation} &\textbf{Test} &  \textbf{mIoU*} & \textbf{mIoU-D*}\\
\textbf{Set} & \textbf{Set} &\textbf{Set} & \textbf{RGB-D} & \textbf{[epochs]}\\
\textbf{} & \textbf{} &\textbf{} & \textbf{[epochs]} & \textbf{}\\
\hline
Semantic & Semantic & Semantic  & 00.92 & 0.82 \\
3D & 3D & 3D   & [100] & [100] \\
\hline
CARLA & CARLA & Semantic  & 0.836 & 0.809 \\
 &  & 3D &  [20] & [20] \\
\hline
CARLA & CARLA & KITTI  & 0.7 & 0.56 \\
 &  &  &   [20] & [20] \\
\hline
% \multicolumn{4}{l}{$^{\mathrm{*}}$Values computed on Test-set.}
\multicolumn{4}{l}{*Values computed on Test-set}
\end{tabular}
\label{tab2}
\end{center}
\end{table}

In Table \ref{tab2} we see that the model trained using CARLA dataset comes close to the model trained on Semantic 3D dataset in only 20 epochs.

\subsubsection{Experiment A}
\begin{figure}
\begin{multicols}{2}
    \includegraphics[width=0.2\textwidth]{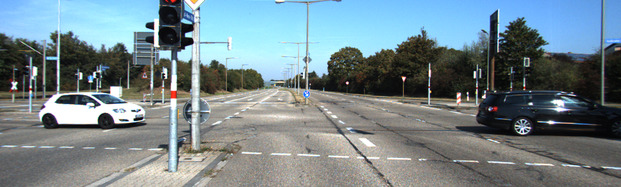}\par 
    \subcaption{}
    \includegraphics[width=\linewidth]{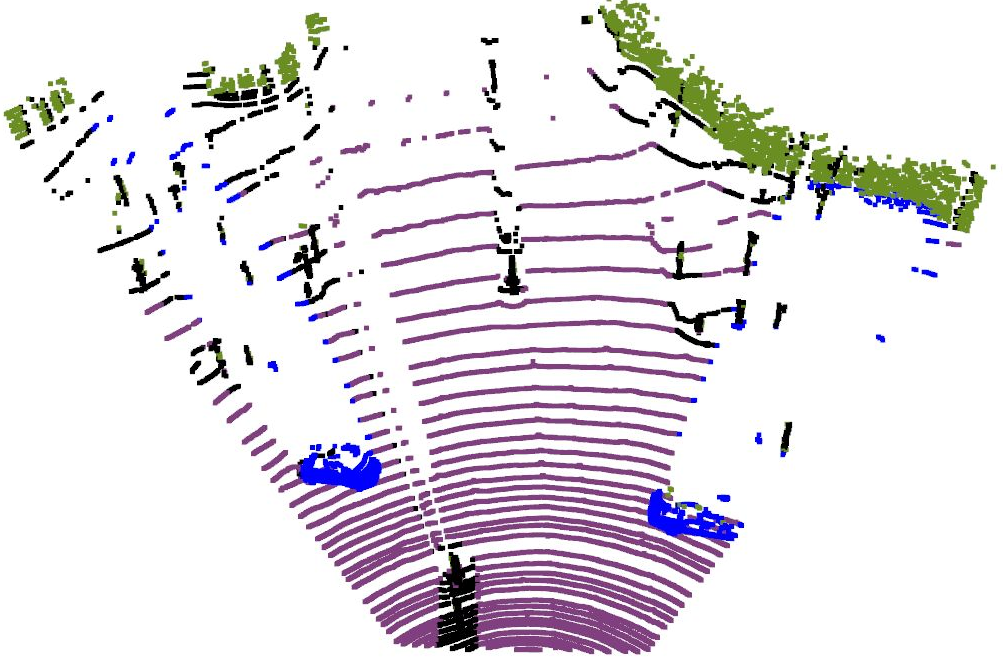}\par 
    \subcaption{}
    \end{multicols}
\begin{multicols}{2}
    \includegraphics[width=\linewidth]{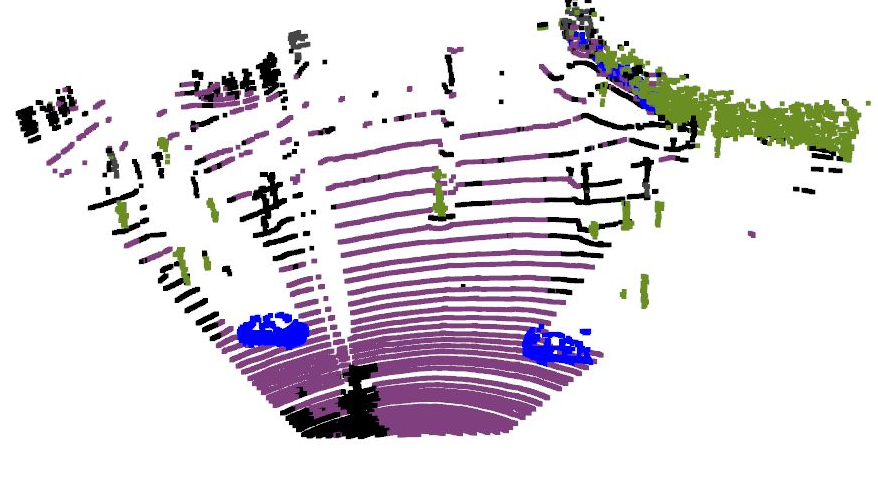}\par
    \subcaption{}
    \includegraphics[width=\linewidth]{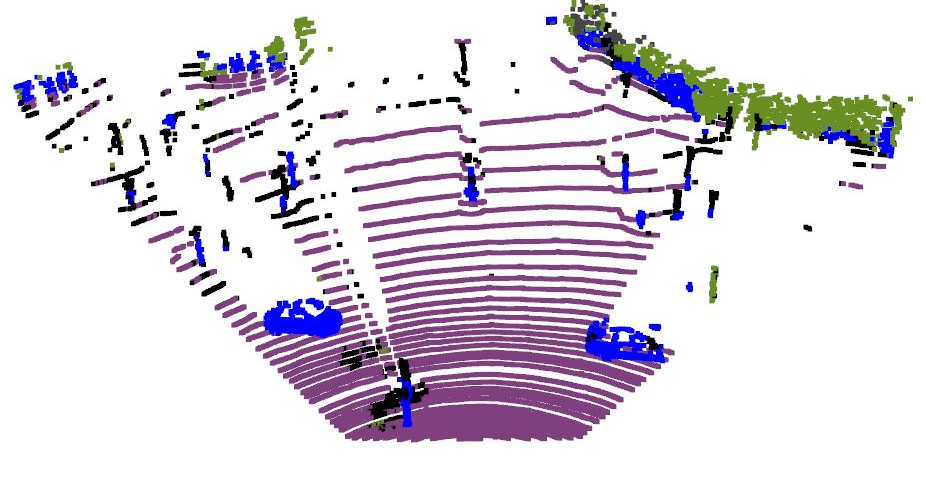}\par
    \subcaption{}
\end{multicols}
\caption{a) Color image of a scene, b) Remapped GT of 19 classes of KITTI  to 4 classes, c) Output of D\textsubscript{4-CARLA}, d) Output of RGB-D\textsubscript{4-CARLA} }
\label{fig:expA}
\end{figure}

In this experiment we show the inference results of RGB-D\textsubscript{4-CARLA} and D\textsubscript{4-CARLA} on a scene from KITTI in Fig. \ref{fig:expA}. We train both RGB-D\textsubscript{4-CARLA} and D\textsubscript{4-CARLA} models after remapping 12 classes of CARLA to 4 common classes as mentioned in column 3 and column 1 of Table \ref{table:sem_class_map_all_to_common4} respectively. While testing on KITTI we remap the 19 semantic classes in KITTI GT images to 4 common classes as mentioned in column 4 and column 1 of Table \ref{table:sem_class_map_all_to_common4} respectively. The resulting GT labels are then reprojected to the corresponding PC and is shown in Fig. \ref{fig:expA}b. By visual inspection we can observe from Fig. \ref{fig:expA}c and Fig. \ref{fig:expA}d that  RGB-D\textsubscript{4-CARLA} has lesser mis-classifications at shorter distances and relatively higher mis-classification at far-away distance  as compared to D\textsubscript{4-CARLA}. For e.g. Vegetation (marked in green color) is labelled as Car (marked in blue). Since there are few points per object at far away distance, the contribution of mis-classification towards the mIOU is not significant. This is also reflected in fourth row of Table \ref{tab2} with a higher mIOU score for RGB-D. 

\subsubsection{Experiment B}

\begin{figure}
\begin{multicols}{2}
    \includegraphics[width=0.2\textwidth]{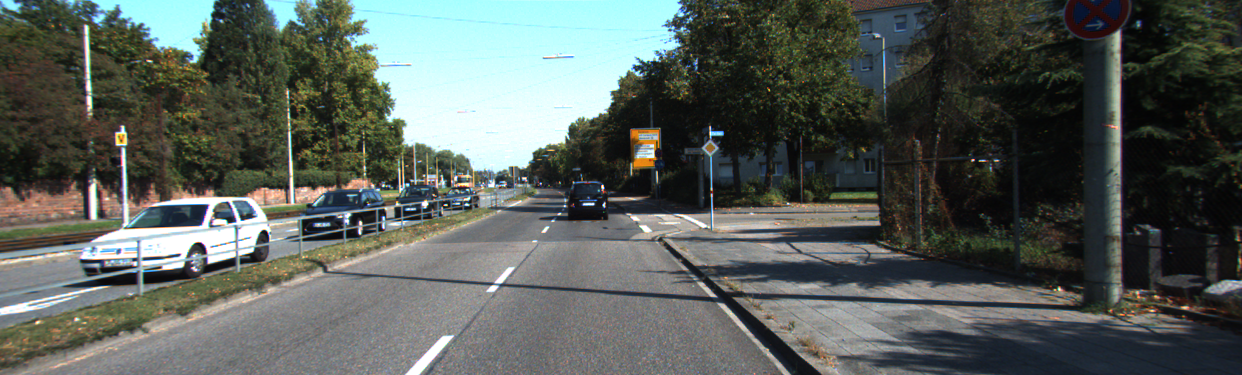}\par 
    \subcaption{}
    \includegraphics[width=\linewidth]{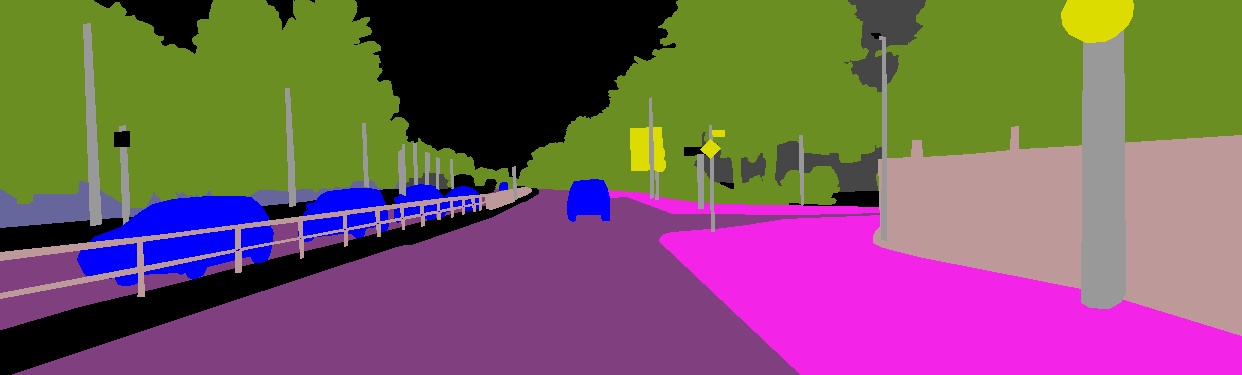}\par 
    \subcaption{}
    \end{multicols}
\begin{multicols}{2}
    \includegraphics[width=0.2\textwidth]{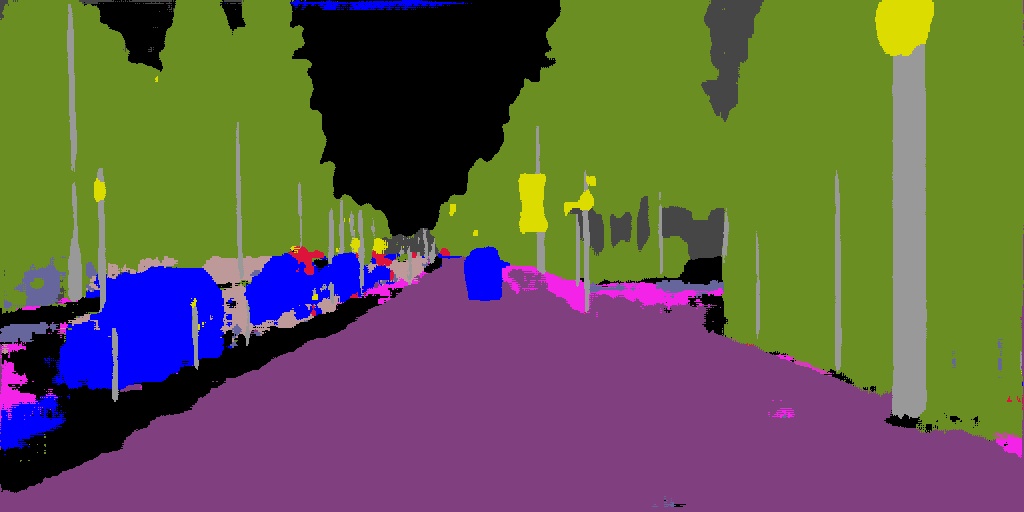}\par 
    \subcaption{}
    \includegraphics[width=\linewidth]{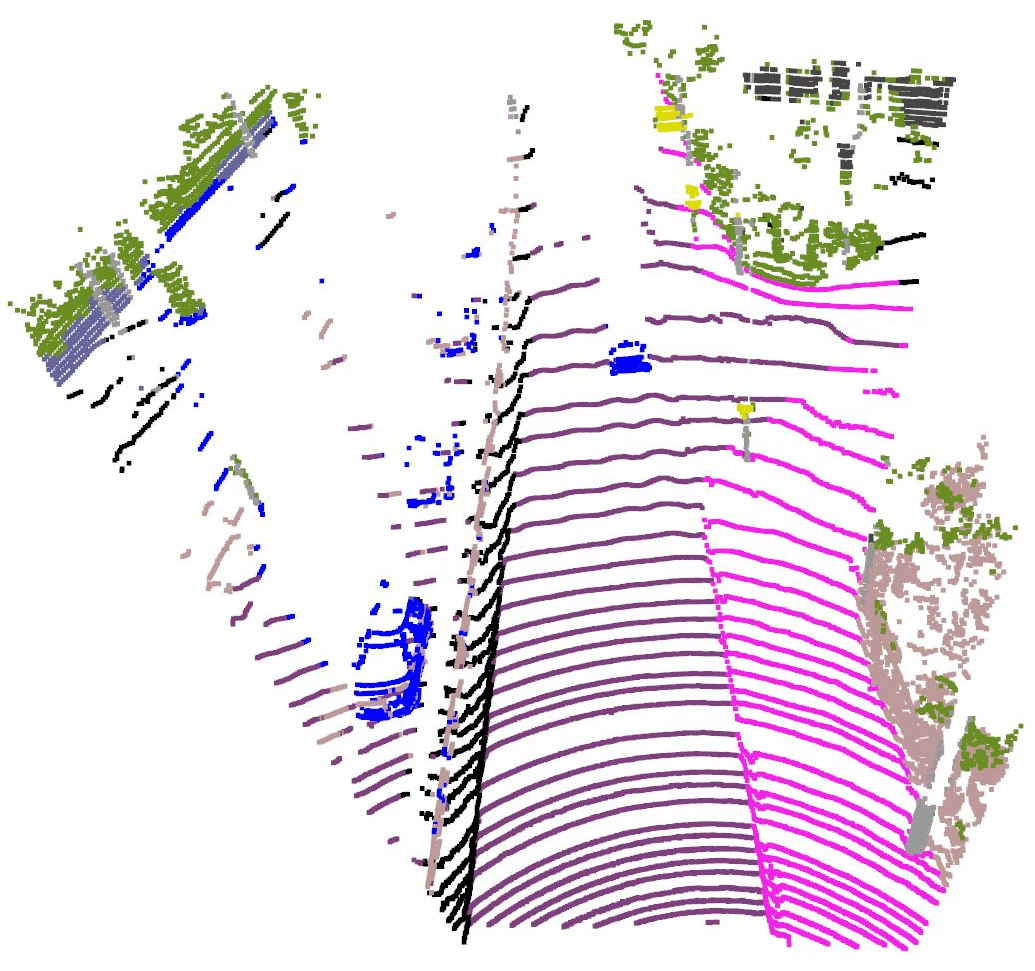}\par 
    \subcaption{}
    \end{multicols}
\begin{multicols}{2}
    
    \includegraphics[width=\linewidth]{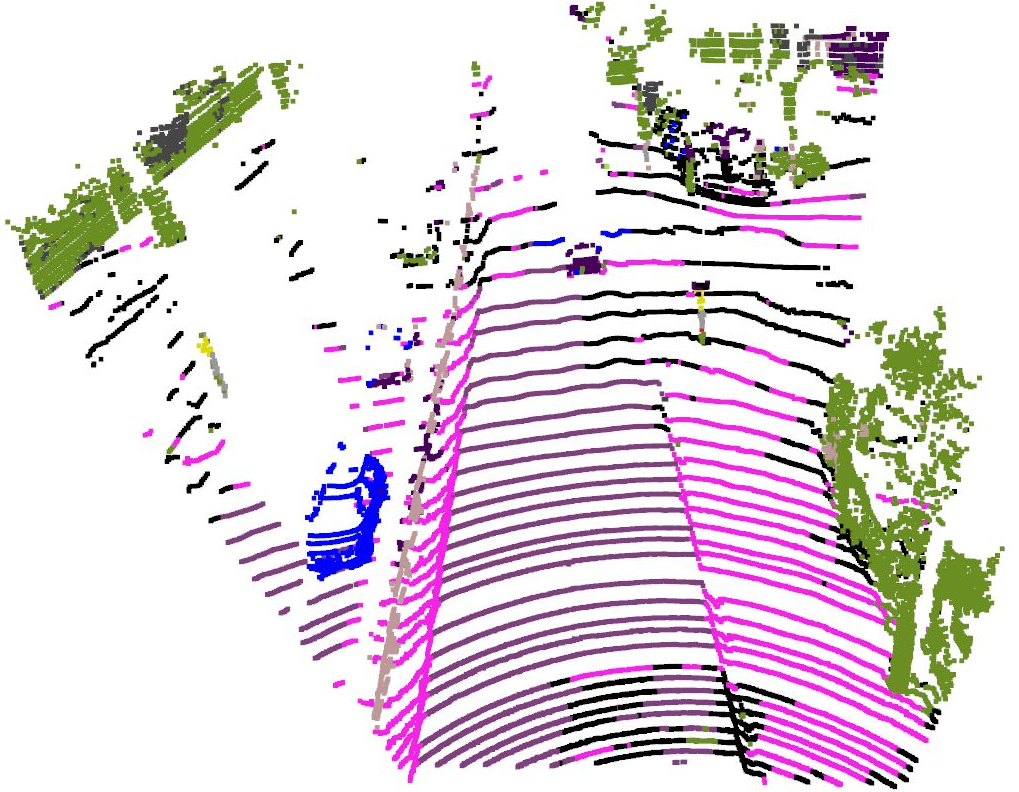}\par
    \subcaption{}
    \includegraphics[width=\linewidth]{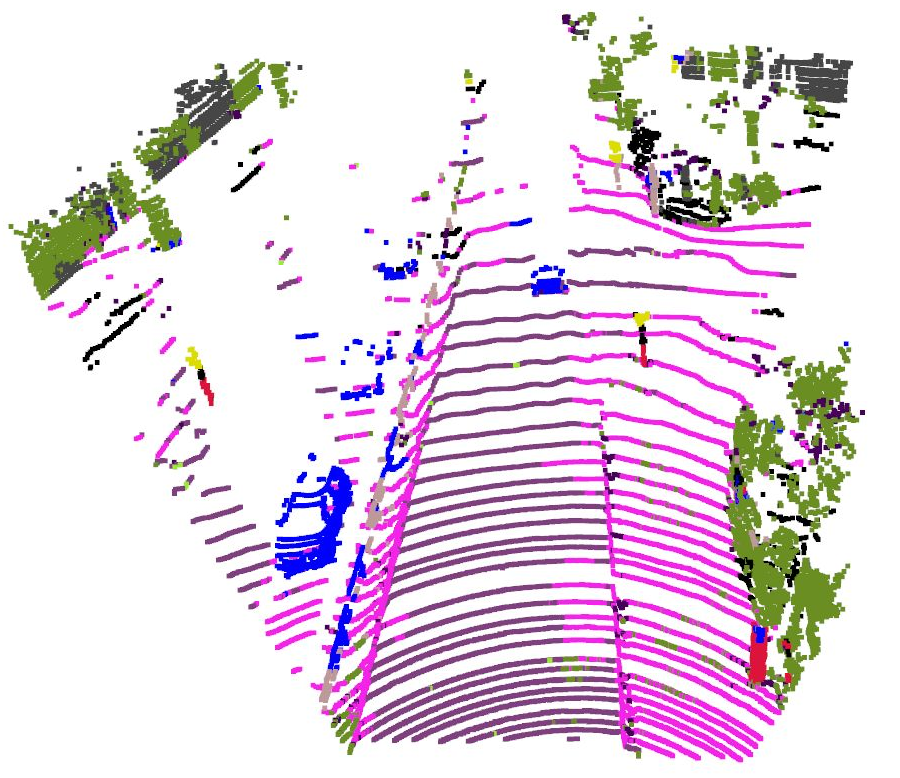}\par
    \subcaption{}
\end{multicols}
\caption{a) Color image of a scene, b) Remapped GT labels of KITTI  to 12 class , c) Output of ERF-Net , d)  Pointcloud representation of b), e) Output of D\textsubscript{12-CARLA}, f) Output of RGB-D\textsubscript{12-CARLA} }
\label{fig:expB}
\end{figure}

To understand the importance of geometric cues we compare our method which uses both geometric and color cues with a 2-D semantic segmentation network that uses only color information. For this purpose we chose pre-trained model of ERF-Net\cite{erf} and fine-tuned it using the GT semantic labelled images of KITTI after remapping the original 19 classes to 12 classes of CARLA, an example of such a remapping is shown in Fig. \ref{fig:expB}b which is GT semantic labels for scene in Fig. \ref{fig:expB}a. We remapped Terrain, Sky, Rider, Truck, Bus, Motorcycle, Bicycle, and Train classes of KITTI to unlabelled and the remaining classes as listed in Fig. \ref{fig:class_dist} were unaltered. The GT for KITTI PC are obtained after re-projecting labels in Fig. \ref{fig:expB}b with corresponding PC and is shown in Fig. \ref{fig:expB}d. The resulting GT PCs are used to evaluate the performance of our multimodal semantic segmentation models. To have a fair comparison we also trained our version of Pointnet++ on CARLA dataset with 12 classes and evaluated the performance on 12 class GT of KITTI PC after remapping. For this we predict all 12 classes using the models trained only on CARLA dataset whereas consider only the 5 major classes of KITTI namely   \textcolor{Building}{Building}, \textcolor{Road}{Road}, \textcolor{Side-walk}{Side-walk}, \textcolor{Vegetation}{Vegetation} and \textcolor{Car}{Car} for computing mIoU as the remaining classes are very less in population. The results are listed in Table \ref{tab3}.

\begin{table}[h!]
\caption{Experimental comparison using All class model}
\begin{center}
\begin{tabular}{|c|c|c|c|c|c|}
\hline
\textbf{Training} & \textbf{Validation} &\textbf{Test} & \textbf{mIoU*} & \textbf{mIoU-D*}\\
\textbf{Set} & \textbf{Set} &\textbf{Set} & \textbf{RGB-D} & \textbf{[epochs]}\\
\textbf{} & \textbf{} &\textbf{} & \textbf{[epochs]} &
% \textbf{Set} & \textbf{Set} &\textbf{Set} & \textbf{classes} & \textbf{[epochs]} &
\textbf{}\\
\hline
CARLA & CARLA & KITTI  & 0.6 & 0.54 \\
 &  &   & [20] & [20] \\
% \hline
% CARLA & CARLA & KITTI  & 0.7 & 0.56 \\
%  &  &  & [20] & [20] \\
\hline
\multicolumn{5}{l}{*On Test set classes - Building, Vegetation, Road, Sidewalk, Cars }
% \multicolumn{4}{l}{Test set are considered}

\end{tabular}
\label{tab3}
\end{center}
\end{table}

It can be seen form Fig. \ref{fig:expB}e and Fig. \ref{fig:expB}f that both RGB-D\textsubscript{12-CARLA} and D\textsubscript{12-CARLA} are able to  segment the side-walk with reasonable accuracy, whereas the inference from ERF-Net (using only the color image in Fig. \ref{fig:expB}a) as shown in Fig. \ref{fig:expB}c mis-classifies side-walk as Road. We speculate the presence of shadows and similar color of road surface and side walk as the main reason. Additionally when we induce a geometric cue from 3D PC, maybe the DNN is able to learn subtle characteristics such as relative difference in elevation (e.g.side-walk is above road), discontinuity in surface normal orientation at the border of two parallel planar surfaces (road and sidewalk) which leads to treat them as two different objects.

The \textcolor{Fence}{Fence} on the extreme right is mis-classified as Vegetation by all models as shown in Fig.\ref{fig:expB}, primarily due to very less representation in the training set.  D\textsubscript{12-CARLA} fails to identify the car at the far end at the middle of the street whereas the RGB-D\textsubscript{12-CARLA} is able to better classify the far away car. This suggests both color and geometric information is necessary for better semantic segmentation of a 3D Point Cloud. 

\section{Conclusion and Future Work}\label{sec:conclusion_and_future_work}
In this work we have shown that it is possible to effectively make use of synthetic data to segment real world 3D Point clouds using multiple modalities namely images and point clouds. In this regard we generated synthetic dataset using CARLA and trained a DNN which is an in-house optimized version of Pointnet++. From our experiments we found that segmentation of static classes such as roads, side-walk, building and objects with rigid geometry such as Cars generalize well when tested on real-world datasets. We also showed that it is possible to obtain better results by training DNN using both color and 3D Point clouds as compared to using either one of them. The methods presented in this work emphasize the role of synthetic datasets especially in cases like segmentation of 3D PC where its difficult to find real-world labelled data and secondly involves huge costs for manual labelling. In such a case the DNN trained on synthetic data can be used to pre-label a large corpus of real-world dataset which potentially would lessen the labelling efforts and thereby significantly reduce labeling cost. 

As part of future work we would like to extend and validate our proposed solution on a much larger corpus of synthetic and real-world datasets.

\end{document}